\documentclass[letterpaper]{article}
\usepackage{amsmath}
\usepackage{aaai19}
\usepackage{times}
\usepackage{helvet}
\usepackage{courier}
\usepackage{url}
\usepackage{graphicx}
\usepackage{hhline}
\usepackage{slashbox}
\title{On the Complexity of Reconnaissance Blind Chess}
\author{Jared Markowitz, Ryan W. Gardner, Ashley J. Llorens \vspace{.2cm}\\
{\large{Johns Hopkins University Applied Physics Laboratory}}\\
\vspace{.1cm}
\scriptsize{\texttt{\{jared.markowitz,ryan.gardner,ashley.llorens\}@jhuapl.edu}}}

\begin{document}

\maketitle
\begin{abstract}
  
This paper provides a complexity analysis for the game of
\emph{reconnaissance blind chess} (RBC), a recently-introduced variant
of chess where each player does not know the positions of the
opponent's pieces a priori but may reveal a subset of them through
chosen, private sensing actions. In contrast to many commonly studied
imperfect information games like poker, an RBC player does not know
what the opponent knows or has chosen to learn, exponentially
expanding the size of the game's information sets (i.e., the number of
possible game states that are consistent with what a player has
observed). Effective RBC sensing and moving strategies must account
for the uncertainty of both players, an essential element of many
real-world decision-making problems. Here we evaluate RBC from a game
theoretic perspective, tracking the proliferation of information sets
from the perspective of selected canonical bot players in tournament
play. We show that, even for effective sensing strategies, the game
sizes of RBC compare to those of Go while the average size of a
player's information set throughout an RBC game is much greater than
that of a player in Heads-up Limit Hold 'Em. We compare these measures
of complexity among different playing algorithms and provide cursory
assessments of the various sensing and moving strategies.

\end{abstract}

\section{Introduction}
Recent successes of artificial intelligence (AI) approaches to strategy games like Go~\cite{Silver2016,Silver2017} and poker~\cite{Brown2017,Moravcik2017} have sparked interest in applying similar techniques to real-world decision-making. The extent to which these superhuman game-playing AI algorithms will generalize to real-world applications is an open question. Autonomous agents applied in practical spaces often encounter imperfect information and a large number of potential actions, sometimes in both continuous and discrete spaces. Further, some decision-making processes require reasoning about both known-unknowns, such as how effective a given strategy may be against a particular adversary, and unknown-unknowns, such as what undiscovered obstacles or threats may lie over the horizon. Each of these can pose serious challenges for today's state-of-the-art AI~\cite{Dietterich2017}. We argue that progress toward robust AI capable of real-world decision-making under uncertainty will require an increased focus on games that better capture the aforementioned challenges. 

Reconnaissance blind chess (RBC) is a recently-introduced variant of chess in which the positions of the opponent's pieces are hidden \emph{a priori} but may be partially revealed through chosen, private sensing actions \cite{Newman2016}. The inherent uncertainty of each player, coupled with the inclusion of the sensing action, provides a unique challenge problem for sensing and resource management algorithms. Historically there have been prior imperfect information variants of popular games such as chess and Go, with corresponding development of game-playing AI algorithms \cite{Russell2005,Cazenave2006,Ciancarini2010}.  Kriegspiel, for example, is a chess variant in which the moves are hidden and the players must decide and act based on incomplete information \cite{Wetherell1972,Li1995}.  In Kriegspiel, partial information is revealed to the players periodically throughout the game by a referee.  RBC may be thought of as a variant of Kriegspiel where the true board state is privately revealed to a given player over a $3 \times 3$ area of their choice prior to each move along with certain properties of the player's own pieces at different points in the game.  We find that the privacy of many of the player's observations significantly increases the uncertainty in the game with respect to the size of information sets within the game, and potentially better aligns it with real-world problems.

While many AI approaches for chess have been developed in recent years (notably Google DeepMinds's AlphaZero algorithm \cite{AlphaZero} and the Stockfish engine \cite{stockfish}), they cannot be successfully applied to RBC without modification since they do not assume or encode uncertainty. Several studies focusing on other imperfect information games published in recent years have begun to address some aspects relevant to RBC.  Neural fictitious self-play (NFSP) is an approach that avoids handcrafted game abstractions and has been applied to two-player zero-sum games with imperfect information \cite{Heinrich2016}. It relies on a combination of intuition gained from games played against experts and refined expertise learned through self-play.  Tree search algorithms in the presence of uncertainty have been evaluated in the context of Kriegspiel \cite{Ciancarini2010} and other games \cite{Silver2010}.  Perhaps most notable have been the recent Libratus and DeepStack results in poker \cite{Brown2017,Moravcik2017}, which leverage counterfactual regret minimization and game-abstraction techniques (e.g., by assuming a king-high flush is roughly equivalent to a queen-high flush or that a bet of \$1000 is roughly equivalent to a bet of \$1001) to find Nash equilibrium strategies for sub-games of reduced complexity \cite{Brown2017}.   RBC does not share this ``local regularity" (e.g., having a rook on one space could have very different implications than having the rook on an immediately adjacent space) and hence does not easily lend itself to such abstraction techniques.  In summary, several recent AI approaches include aspects that could be useful in developing strong RBC playing algorithms; however none is immediately applicable to RBC.

In this paper, we leverage commonly-used measures of game complexity to show that RBC combines some of the most challenging aspects of popular perfect- and imperfect-information games. These measures include information set size as well as the number of possible information sets encountered at each action throughout a game.  By simulating tournaments among different canonical reconnaissance chess playing algorithms, we illustrate the influence of different sensing and moving strategies on the game complexity imposed on each player.  We demonstrate that the number of possible information sets encountered throughout a game compares with that of Go, even for effective sensing strategies.  We further find that the average size of a player's information set throughout a game is comparable to that of Two-Player Limit Texas Hold 'Em.


\section{Reconnaissance Blind Chess}
Reconnaissance Blind Chess (RBC)~\cite{Newman2016} is distinct from standard chess in that a player does not know \emph{a priori} where the opponent's pieces are.  Instead, each player is allowed to sense a $3 \times 3$ grid of their choosing before each move they make.  An arbiter returns the ground truth contents of this $3 \times 3$ grid, which the player then uses to inform his or her next move.  The arbiter also notifies a moving player of the result of each move including where her piece lands and whether a capture was made.  The arbiter notifies the non-moving player when one of her pieces has been captured, specifying the location of the captured piece.

To accommodate these limited sources of information, the rules of RBC diverge from standard chess in a few additional ways.  The notions of check and checkmate are eliminated, with the game ending only when one player captures the other player's king (or when time runs out if playing with a clock).  Moves that put a player into check are now allowed, as one or both of the players may not be aware that check has been reached.  On a given turn, a player may command a move that is illegal on the true game board (either intentionally or accidentally).  If the move involves a sliding piece that is blocked by one or more opponent pieces, the moving piece captures the first piece in its path and remains at the location of the captured piece.  Illegal castles are not included in this; if an opponent's piece blocks an attempted castle, no pieces are moved and the turn passes over to the opponent.  Similarly, if the commanded move is a forward or diagonal pawn move (including \emph{en passant}) where an opponent piece is or is not present, respectively, nothing happens except that the turn passes over to the other player.\footnote{If a pawn attempts to move forward 2 squares and only the second square is blocked, the pawn will move forward 1 square.}  Finally, a player may explicitly command a ``pass" move in RBC.  This is not legal in normal chess but can provide an advantage in RBC by increasing the opponent's uncertainty about the board state (by not allowing them to verify the move through sensing).

The uncertainty induced by the rules of RBC force each player to encounter non-singleton information sets throughout the game.  In any play sequence, both the size of a player's current information set and the number of information sets reachable by a given action depend strongly on the strategy of each player.  We therefore chose to study the complexity of reconnaissance blind chess through the use of Monte Carlo methods involving games among several diversely configured bot players with conceptually-simple playing algorithms.  Note that these algorithms are intended to help us characterize complexity relative to fairly intuitive, straight-forward approaches and provide basic insights, not to approach optimal strategies.

\section{Playing Algorithms}
We chose five different bots with which to evaluate the complexity of RBC.  These bots represent a variety of basic playing strategies and were additionally used to glean basic insights on the effectiveness of different tactics.  For each bot combination, we played 50 games: 25 with a given player as white and 25 with that player as black.  For the purposes of this analysis, the ground truth board was made available to two bots (RandomBotX and PerfectInfoBot).  Multiple bots made use of a deterministic flavor of the Stockfish engine \cite{stockfish}, which is currently the strongest computer chess program that is publicly available.  The complete list of bots used in this analysis, the rationales behind them, and their implementation specifics are given below.

\subsection{RandomBotX}
RandomBotX both senses and moves (or passes) randomly.  The ``X" in the name of this bot refers to the percentage of the time the bot chooses to pass, which was fixed at 25\% unless otherwise indicated.  On every turn where the bot does not pass, the bot chooses randomly from the legal RBC moves available on the ground truth board.   For the sensing operation, RandomBotX chooses with uniform probability among the 36 sensing areas where the entire $3 \times 3$ area is on the board.  RandomBotX was chosen as a baseline, essentially as a ``worst case" for accumulating and imposing uncertainty on the opponent.  It rarely senses the true move of an opponent and its moves cannot be consistently predicted (or sensed).

\subsection{NaiveBot}
The NaiveBot performs in a rapid but naive fashion.  For sensing, it chooses the $3 \times 3$ area whose squares have, on average, gone longest without being sensed.  This sensor only considers the 36 sensing areas where the entire $3 \times 3$ area is on the board while considering the positions of its own pieces as having gone 0 turns without being sensed. The NaiveBot maintains a simple, single hypothesis of the opponent's state where only the $3 \times 3$ sensed region is directly updated to match the results of each sense and other squares are left unchanged with two exceptions. First, uniquely-identifiable pieces, namely the king, queen, white bishop, and black bishop are removed from their old position if sensed in a new position.  Second, the king is replaced near its previously known position in a location that is consistent with the sense results if a sense would result in king removal. NaiveBot moves by applying the top move as recommended by Stockfish on the current (generally flawed) hypothesis.  NaiveBot was chosen to see how a simple-minded and fast yet reasonable combination of sensing, estimating, and moving strategies would perform against other baseline bots.

\subsection{MHTBot}
The multi-hypothesis tracking (MHT) bot computes and stores all possible boards throughout the game and uses the current set to choose its senses and moves.  For sensing, MHTBot chooses the sense location that minimizes the expected number of possible boards on the next turn, assuming that each currently possible board is equally likely.  The bot moves by choosing the mode best move selected by Stockfish over all possible boards.  The MHTBot was chosen to include both strong sensing and moving strategies that require no assumptions about the opponent's strategy.

\subsection{PredictorBot}
Like the MHTBot, the PredictorBot stores all possible boards throughout the game and moves by choosing the mode of the best move selected by Stockfish over all possible boards.  It differs in its sensing strategy, which attempts to predict where the opponent will move in a given turn.  To accomplish this, PredictorBot computes a weight for each square of the board, which is initialized to 0.  It randomly selects up to 512 boards from the set of possible boards its opponent could have encountered on the last move and computes the top 5 Stockfish moves on each selected board, generating 5 destination squares and scores.  The scores are converted to weights by subtracting out the minimum score and normalizing.  Each weight is then added to the weight maintained for the destination square of the corresponding move.  The final sensing location is chosen to be the one where the sum of the weights of the sensed squares is maximal.  This bot was chosen to illustrate the impact of a basic attempt to predict the moves of the opponent or to increase the relative importance of being aware of classically stronger moves as compared to classically weaker ones.

\subsection{PerfectInfoBot}
This bot is given access to the ground truth board and always chooses the best move on that board according to Stockfish.  It is included as an example of an extremely strong but predictable bot and as a means to evaluate our metrics for a bot that always knows the full position of its opponent.

All of the bots listed above break ties by choosing from the best available options in a given situation uniformly at random.  The standard threefold repetition and fifty-turn draw conditions of chess were enacted in games that include the PerfectInfoBot but not in others, as bots with imperfect information cannot properly evaluate the truth board to declare a tie.  The addition of these conditions for the PerfectInfoBot was found to be necessary because of the infinite loops that can otherwise occur when two identical PerfectInfoBots (with the same moving strategy) face off.

\section{Game Complexity}
We focus on measuring game complexity in two ways: through the size of a game and through the degree of uncertainty in a perceived state, or the number of true states that are possible in a given perceived state.

\begin{table*}[tbh]
  \centering
  \begin{tabular}{c|c|c|c|c}
Lim 2-P Poker & Chess & Recon Chess & Lrg No-Lim Poker & Go (19x19) \\
\hhline{=|=|=|=|=}
$10^{13}$ & $10^{43}$ & $10^{139}$ & $10^{162}$ & $10^{170}$ \\
\end{tabular}
\caption{The approximate size of several games in terms of the number of information sets for imperfect information games and the number of distinct states in perfect information games.}
\label{tab:game_size}
\end{table*}

\subsection{Size of the Game}
We measure the size of imperfect information games in terms of possible information sets.  Informally, an information set is the state of the game from a given player's perspective, given what that player has observed.  In a perfect information game, the size of a player's information set is always 1.  In an imperfect information game, the size of the information set can grow or shrink throughout the game as it accounts for all possible configurations of hidden information (including the opponent's knowledge).

The size of a given game can be measured using the number of possible information sets that can be encountered by a given player in that game.  The true size of a game is a fixed quantity but is often intractable to compute.  It has not been computed exactly for standard chess, partly because of the varying number of different moves available at different junctures and partly because of the variance in the total length of the game.  The true size of reconnaissance blind chess would be even more difficult to compute, due to the imperfect information and sensing action.  Hence we chose to compute an approximation of the practical size of RBC, using the same general approach that was used by Claude Shannon to compute the game tree size for standard chess \cite{Shannon1950}.  This ``Shannon number" was computed based on both a typical branching factor and a typical game length, multiplying the same branching factor repeatedly for each turn for each player.  For RBC, we estimated the game size by evaluating a series of games between two MHTBots.  We chose the MHTBot because it explicitly minimizes the expectation of the size of the information set after each sensing operation, making no assumptions about the opponent's moving strategy.  The MHTBot is thus both conservative and generic in the size estimates it provides.  Each RBC game size was computed by multiplying the number of distinct information sets possible after each action throughout the game, similar to what Shannon did with chess but this time including the sensing action.

We report the mean of this RBC game size metric across all simulated games.  Our results are given in Table \ref{tab:game_size}, along with estimates or exact size measurements for other common games.  For perfect information games, the table gives the size of the game with respect to the number of distinct states in the game.  For imperfect information games, it presents the size in terms of information sets. Lim 2-P Poker refers to Limit Heads-Up (i.e., 2-player) Texas Hold 'Em and Lrg No-Lim Poker refers to Texas Hold' Em with \$50-\$100 blinds, \$20,000 stacks, and where bets can be made with \$1 granularity. These exact sizes in terms of information sets are from Johanson \cite{Johanson2013}.  The estimate for the number of distinct states in Go with a $19 \times 19$ board is from Tromp \cite{Tromp2016}.   The estimated number of states for chess comes from Shannon \cite{Shannon1950}.  For reference, the number of atoms in the observable universe is approximately $10^{80}$ \cite{Villanueva2009}.

\begin{table*}[tbh]
 \centering
  \begin{tabular}{l||c|c|c|c|c}
   Included in the \emph{State} & 2-P Poker & Chess & RBC w/ MHTBot & 6-P Poker & Go (19x19) \\
  \hhline{=#=|=|=|=|=}
      concrete pieces/cards & $1,083$ & $1$ & $181$ & $6.4*10^{14}$ & $1$ \\
      above + opponent's knowledge & $1,083$ & $1$ & $1.3*10^{68}$ & $6.4*10^{14}$ & $1$ \\
\end{tabular}
\caption{The approximate mean number of possible opponent states through the game in a given information set (perceived state), for several games.}
\label{tab:info_set_size}
\end{table*}

\subsection{Degree of Uncertainty in a Perceived State}
\label{sec:uncertainty_in_state}
Another metric used to measure imperfect information games is the number of possible states in the information set.  This metric elucidates how difficult it is to evaluate the quality of a given information set, i.e., how difficult it is to evaluate a given perceived state.  Note that this number of possible states could be measured in several ways.  One way would be with respect to the number of possible concrete states in the game's hidden information at a given point in time. In RBC, this would count the number of possible ways an opponent's pieces can be arranged (including what castling rights they have and where \emph{en passant} capture may be possible).  A second way the size of an information set could be measured is to include the opponent's possible knowledge as part of the state.  For example, in RBC this would consider all series of sense possibilities that would lead to different opponent knowledge to be different states in the information set.  Realistically, the opponent's knowledge is essential information to creating an optimal strategy. To illustrate this, consider the extremes.  If one assumes the opponent knows our exact state, we could never risk sneaking up on the opponent.  If one assumes the opponent knows nothing of our state, we could make overly risky moves.

For the majority of commonly played and studied games (such as Texas Hold 'Em), these two measurements are equivalent.  This is because the games are frequently set up to have information that is public to all players and information that is private to individual players; the players know that the private information is private and the public information is public.  The games typically do not have a setup where an opponent can learn some information about another player's state without that player knowing.\footnote{This is actually challenging or potentially inconvenient to set up in games without some type of arbiter, which includes most physical games, but is simple to do with an arbiter in an electronic setup like that of RBC.}  This uncertainty about the opponent's knowledge is a key property of RBC that differentiates it from other games.  The opponent has 36 different sensing options each turn, which can all potentially lead to different information, and one never learns with certainty where the opponent has sensed.  Thus we can roughly estimate the expansion factor in the number of states in a given information set to be $36^n$, where n is the number of turns that have taken place so far.  This does not include information that the opponent may learn from the results of his or her moves.  This exponential expansion in the size of each information set with no ability to reduce the size with certainty makes successful implementation of algorithms like counterfactual regret minimization \cite{Zinkevich2008} (even online, Monte Carlo approaches \cite{Lisy2015}) extremely difficult because the approaches would involve sampling the possible series of the opponent's senses.  However, having this uncertainty in the opponent's knowledge seems critical for adversarial real-world applications where one would rarely know exactly what one's opponent knows.
 
To complement our rough estimate of $36^n$ in the expansion of the number of states in an information set due to different opponent knowledge, we also computed the size of each current information set with respect to the first definition we provide.  That is, we computed the number of possible opponent states with respect to explicit game information as well as the state of the opponent's pieces, through a series of games.  We report the mean number of states of the opponent's pieces before the sensing action in the first data-row of Table \ref{tab:info_set_size} across all turns and all games for the MHTBot.  $N$-P Poker refers to Texas Hold 'Em with $N$ players.\footnote{We computed this as $\left( \binom{50}{2} + \binom{47}{2} + \binom{46}{2} + \binom{45}{2} \right) /4$ for 2-player poker to get an average number of possible states before flop, before the turn, before the river, and after the river.  We used an analogous computation for 6-player poker.}  In the second row of the table, we provide the estimated number of states accounting for the opponent's knowledge, i.e., for RBC we add an additional factor of $36^{n-1}$ for the size of the information set each turn (where n is the turn number).

As can be seen from the table, when only considering the positions of the pieces, the number of states in an information set can actually be kept relatively low (an average of less than 200) using a bot (the MHTBot) whose sensing strategy is to minimize the expectation value of that number of states.  However, when we account for the different possible states of the opponent's knowledge, this set becomes much larger.  On the surface it appears to be significantly larger than that of poker.  However practically speaking it may be much larger, which we discuss further below. 

\subsection{Impact of Different Strategies on Complexity}
Critical to a bot's performance in RBC is its ability to both manage its own uncertainty and to impose uncertainty on its opponent.  We evaluated the abilities of the bots in our canonical set in these areas, again through the use of information sets.

\begin{figure} 
\includegraphics[width=1.0\linewidth]{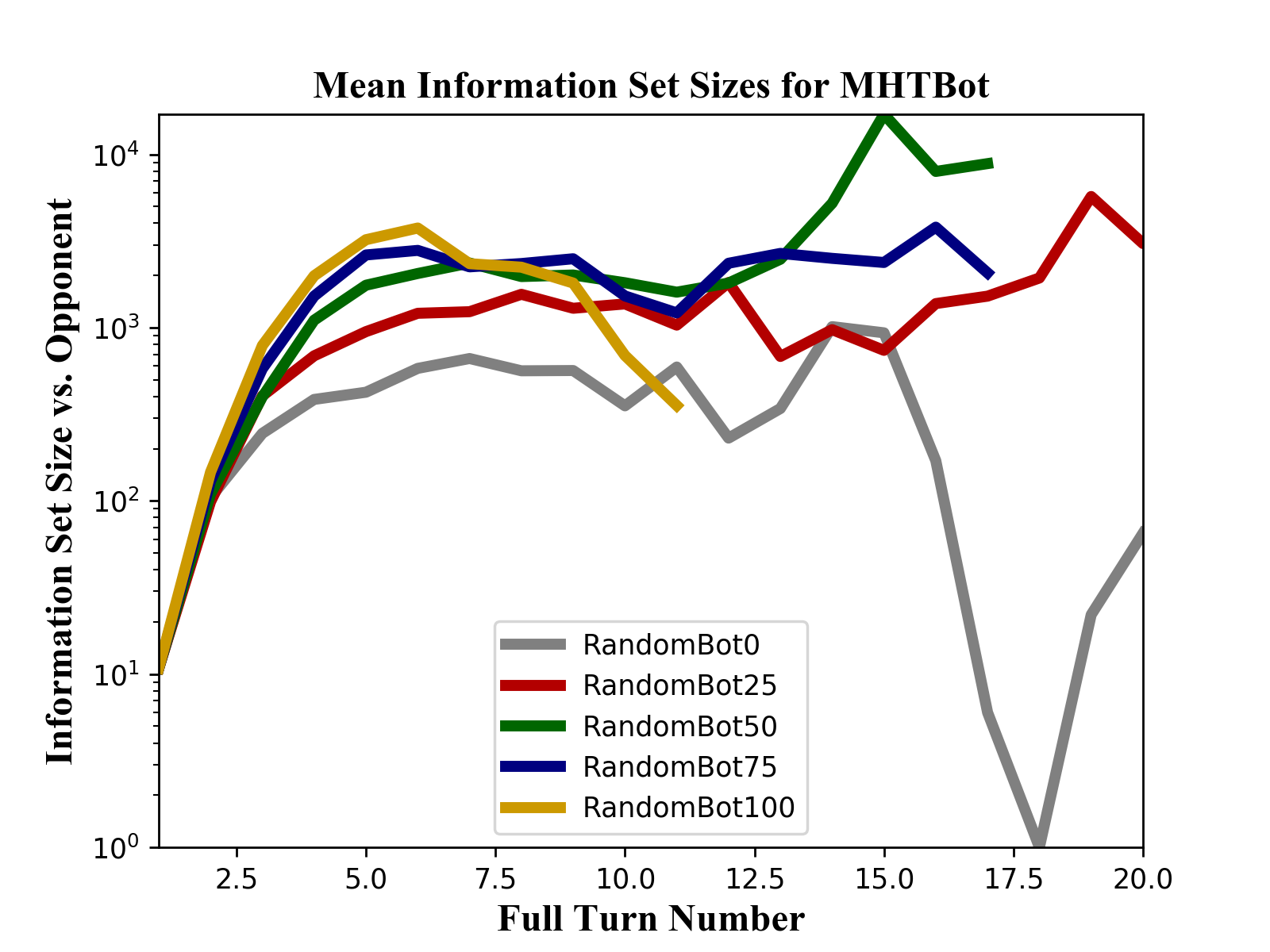}
\caption{The mean information set size observed prior to sense by MHTBot, plotted as a function of turn throughout a series of games against RandomBotXs with different pass probabilities.}
\label{fig:info_set_size_mht_random}
\end{figure}

\begin{table*}[tbh]
\centering                 
  \begin{tabular}{l||c|c|c|c|c|}
  \backslashbox{\textbf{white}}{\textbf{black}}  & RandomBot25 & NaiveBot & MHTBot & PredictorBot & PerfectInfoBot \\                 
    \hhline{=#=|=|=|=|=|}
    RandomBot25 & 10-15-0 & 0-25-0 & 0-25-0 & 0-25-0 & 0-25-0 \\
    \hline
    NaiveBot & 25-0-0 & 9-16-0 & 4-21-0 & 6-19-0 & 0-25-0 \\
    \hline
    MHTBot & 25-0-0 & 22-3-0 & 15-10-0 & 3-22-0 & 0-25-0 \\
    \hline
    PredictorBot & 25-0-0 & 22-3-0 & 23-2-0 & 13-12-0 & 0-0-25 \\
    \hline
    PerfectInfoBot & 25-0-0 & 25-0-0 & 25-0-0 & 25-0-0 & 0-0-25 \\
    \hline                                        
\end{tabular}
\caption{Color-dependent win-loss-draw numbers for each bot against the others.}
\label{tab:records}
\end{table*}

The first area we explored was the effect on the opponent's information set of simply doing nothing.  To accomplish this, we played a set of RandomBotXs against our MHTBot.  These RandomBotXs differed in their pass frequency, ranging from no passes commanded to a bot that passed all of the time.  The sizes of the information sets observed by the MHTBot against these bots are plotted in Figure \ref{fig:info_set_size_mht_random}.  We observe that increasing the amount of passing generally increases the size of the opponent's information set (although it may prevent the player in question from making strategic progress).  This effect occurs because of the inability of the opponent to (almost ever) determine with certainty that a pass move has occurred.  The pass move introduced in reconnaissance blind chess may thus play an important strategic role, perhaps mimicking the benefits of ``waiting out" your adversary in a real-world situation.  With this experiment in hand, we decided to use RandomBot25 (i.e. a RandomBotX that passes 25\% of the time) in all subsequent experiments.
 
We conducted a full round-robin tournament among our five core bots: RandomBot25, NaiveBot, MHTBot, PredictorBot, and PerfectInfoBot.  The color-dependent win-loss-draw numbers for each bot paired with each other bot are presented in Table \ref{tab:records}.         

We observe that there is a fairly clear hierarchy of bots with RandomBot25 always losing and PerfectInfoBot always winning or tying.
The results illustrate the nature of imperfect-information games, where even seemingly inferior strategies win some of the time (as can be seen in the NaiveBot vs. MHTBot match-up, for example).  We notice that the PredictorBot fares reasonably well against the PerfectInfoBot.  This happens because the PredictorBot's prediction strategy assumes the same algorithm as the PerfectInfoBot's playing strategy (deterministic Stockfish) and thus PredictorBot typically senses in a way that identifies the PerfectInfoBot's move.  This highlights a known game-theoretic result: even with perfect information/sensing an RBC bot should mix its strategies in order to avoid being transparent to a bot customized to its strategy.

\begin{figure} 
\includegraphics[width=1.0\linewidth]{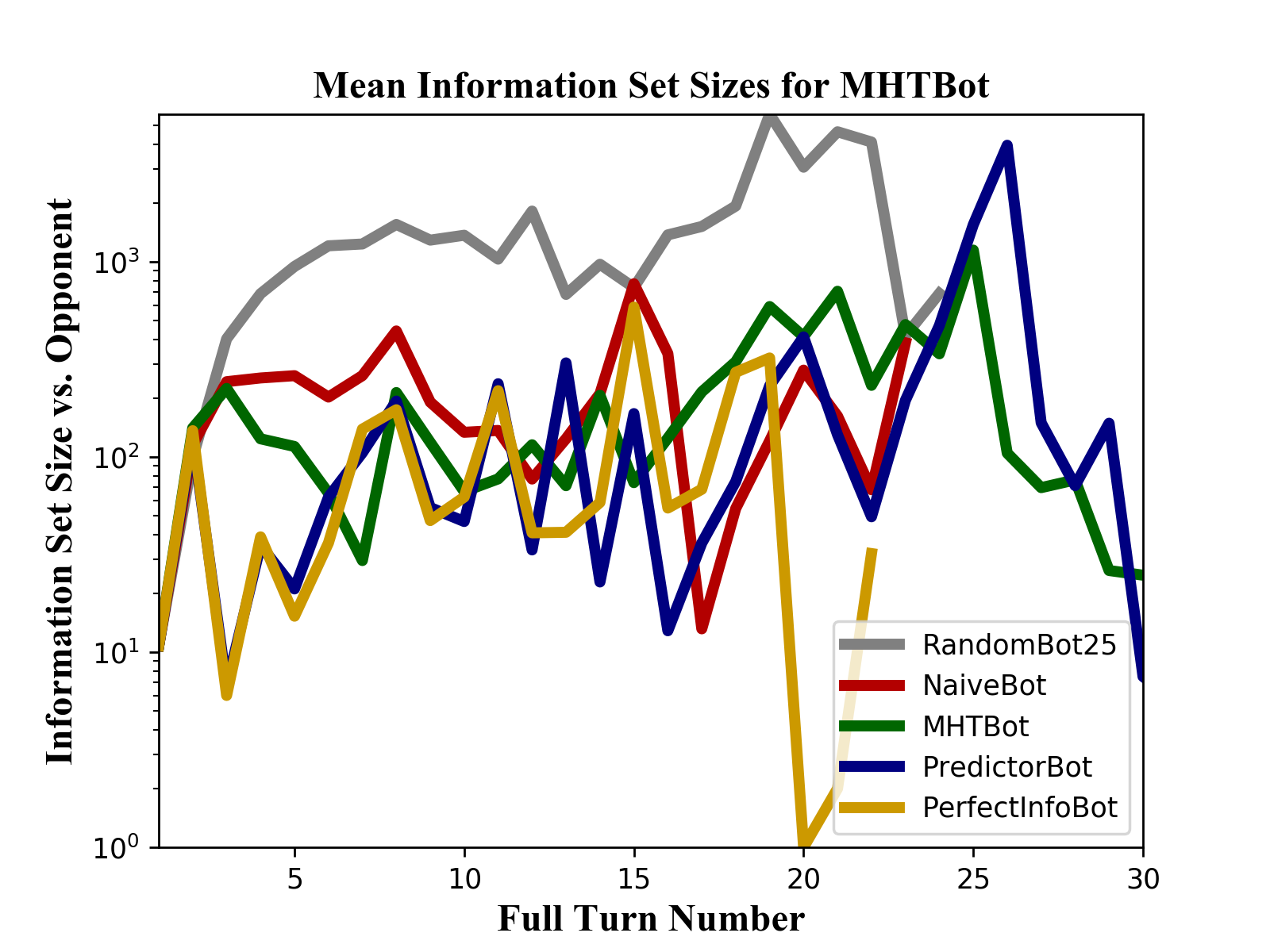}
\caption{The mean information set size observed prior to sense by MHTBot, plotted as a function of turn throughout a series of games against the various bots chosen for this analysis.}
\label{fig:info_set_size_mht_all}
\end{figure}

\begin{figure} 
\includegraphics[width=1.0\linewidth]{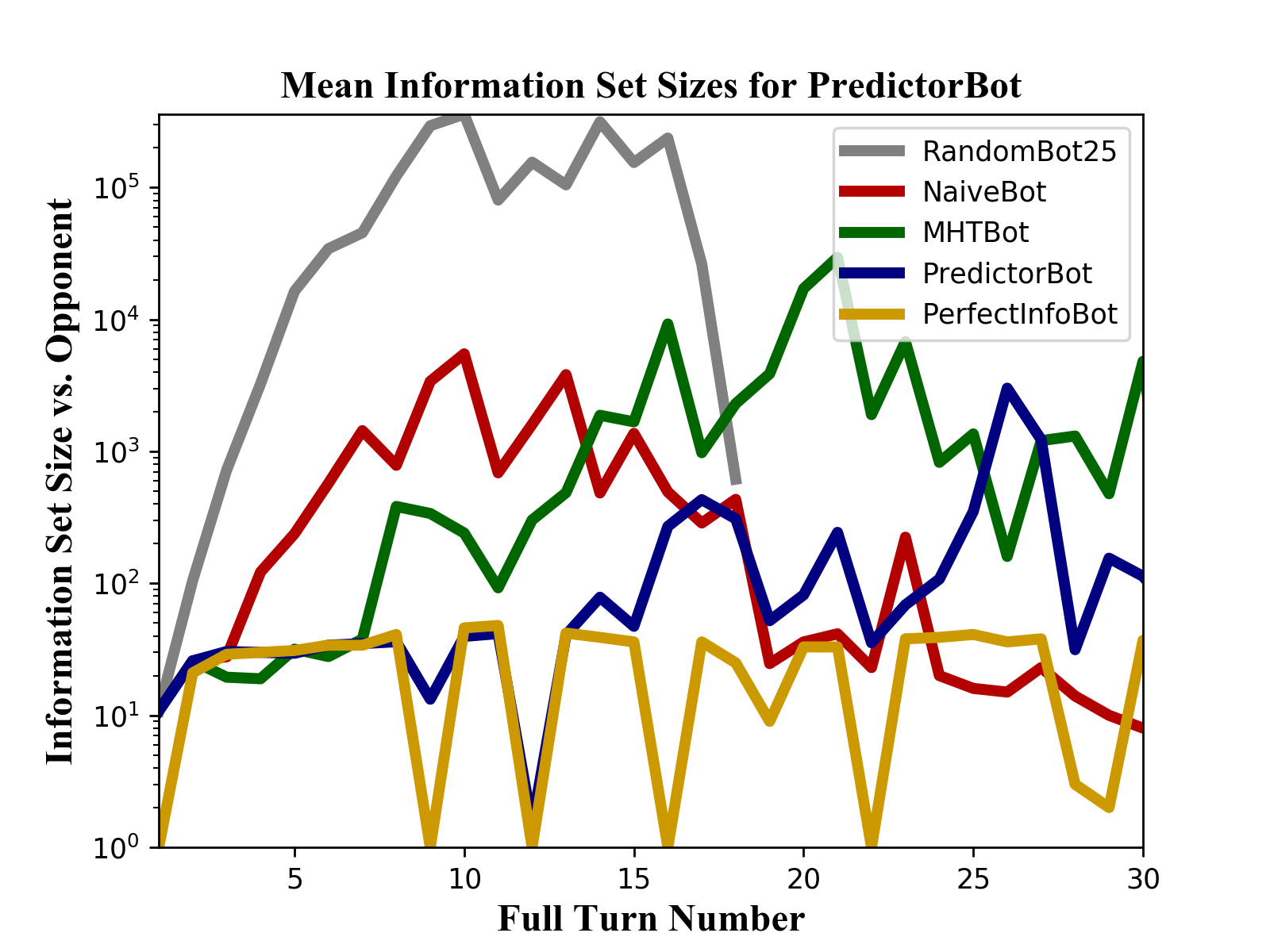}
\caption{The mean information set size observed prior to sense by PredictorBot, plotted as a function of turn throughout a series of games against the various bots chosen for this analysis.}
\label{fig:info_set_size_predictor_all}
\end{figure}

Throughout our bot tournament, we tracked both the size of the information set observed by each bot and the number of potential information sets possible (i.e. the information set branching factor) as a result of each action.  The size of the information sets encountered by the MHT bot and the PredictorBot are shown in Figures \ref{fig:info_set_size_mht_all} and \ref{fig:info_set_size_predictor_all}.  The MHT bot represents a principled way of minimizing the size of the information sets observed without making any assumption about the opponent's strategy.  The PredictorBot, on the other hand, tries to minimize the size of its information set by guessing the moves of its opponent (in this case by assuming that the opponent moves similar to how Stockfish would move).  As Figures \ref{fig:info_set_size_mht_all} and \ref{fig:info_set_size_predictor_all} illustrate, compared to MHTBot the PredictorBot is very effective in minimizing its information set size against PerfectInfoBot, is slightly better at minimizing its information set size against itself, and is significantly worse at minimizing the size of the information set against every other opponent.  These results essentially correspond to how closely the opponent bot replicates the Stockfish mover (including its strategy and available information).

Finally, we display the average information set branching factor per turn (sense and move) in Table \ref{tab:branching_factor}.  These results demonstrate the importance of the bot sensor.  The RandomBotX result shows how dangerous a lack of coherent sensing strategy can be; in fact, we were only able to compute its information set sizes and branching factors through 6 turns due to memory constraints.  On the other hand, the random move strategy employed by RandomBotX is seen to impose significantly more uncertainty on its opponents.  The sensing approach of NaiveBot shows some utility, but is not nearly as adept at reducing uncertainty as the sensing approach used by MHTBot.  The MHTBot method is by far the most consistent of the approaches used by our bots in terms of both the size of the information sets produced and their branching factor.  The PredictorBot does better against itself and PerfectInfoBot, but significantly worse elsewhere.  The PerfectInfoBot may not be fairly compared with the others, but we include its information set branching factor (caused by the moves of the player and its opponent) for completeness.
 
 \begin{table*}[tbh]
 \centering
  \begin{tabular}{l||c|c|c|c|c|}
  \backslashbox{\textbf{player}}{\textbf{opponent}}  & RandomBot25 & NaiveBot & MHTBot & PredictorBot & PerfectInfoBot \\             
    \hhline{=#=|=|=|=|=|}
    RandomBot25 & 160,428 & 60,517 & 51,955 & 83,174 & 73,412\\
    \hline
    NaiveBot & 168,647 & 55,606 & 13,793 & 11,803 & 11,806 \\
    \hline
    MHTBot & 27,962 & 10,100 & 7,619 & 6,592 & 6,541 \\
    \hline
    PredictorBot & 444,431 & 20,569 & 28,618 & 3,448 & 2,063\\
    \hline
    PerfectInfoBot & 40 & 39 & 39 & 35 & 33 \\
    \hline
    \end{tabular}
\caption{The mean information set branching factor for each type of bot against each opponent.}
\label{tab:branching_factor}
\end{table*}          

\section{Discussion}
We have measured the complexity of RBC in terms of both the approximate size of the game and the approximate size of a perceived state.  However, many more qualitative factors impact the practical complexity of a game beyond these measurements.  We discuss these factors with relation to recent successes in poker and Go, making additional observations about RBC as shown from our data. 

\subsection{State Abstraction}
One important property that affects game complexity relates to a player's ability to create abstractions of various possible states.  In the large no-limit poker game referenced in Table \ref{tab:game_size}, for example, a bet of \$10,000 is very similar to a bet of \$10,001, and probably even close to bets of \$11,000 or \$12,000.  One could potentially abstract these bets to have the same meaning.  If they were considered equivalent, the number of information sets in the game would drop drastically.  Similarly, a hand with a 4 and a 9 may be very similar to a hand with a 6 and a 9.  Brown and Sandholm were able to take advantage of such abstractions in the creation of Libratus \cite{Brown2017}.  However, chess and RBC yield no obvious analog.  Having a piece in one square compared to an adjacent square can significantly change the game, for example.  Without such abstraction techniques, the number of information sets presents a practical challenge for directly applying algorithms like counterfactual regret minimization \cite{Zinkevich2008}, which was leveraged by Libratus.

\subsection{State Evaluation}
One of the major challenges in creating strong algorithms for Go is the difficulty of evaluating the strength of a player's current position.  In chess, however, there are simple (albeit potentially flawed) ways to heuristically estimate if a player has an advantage.  For example, one can assign a value to each piece a player has and to each piece her opponent has.  If a player is up a knight, for instance, that indicates she has an advantage.  Other heuristics are present in chess but are largely absent in Go.  Ultimately DeepMind was able to overcome this challenge in AlphaGo, Alpha Go Zero, and AlphaZero \cite{Silver2016,Silver2017,AlphaZero} (all stronger than the human world Go champion) through the combination of two techniques.  The first was Monte Carlo tree search (MCTS; \cite{Browne2012}), which utilizes random playouts down the game tree in order to better evaluate a player's current options.  The second was the use of deep neural networks trained to provide a novel state-evaluation function that efficiently guided the tree search.

The difficulty of state evaluation in RBC may actually be more similar to Go than to chess. With the uncertainty of RBC, for example, a player does not necessarily know even what pieces her opponent has.  This makes state evaluation exceedingly difficult.  Deep neural networks (as used in AlphaGo) may once again prove useful here, this time for evaluating \emph{information sets} and once again for guiding searches.  However pure MCTS is not applicable, as it relies on perfect information.  Imperfect information variants of MCTS exist, but their success is not ensured.  Information set Monte Carlo tree search (ISMCTS) \cite{Cowling2012} is intended for games of imperfect information, but it is not guaranteed to converge to an optimal strategy (even given infinite time) due to the locality problem~\cite{Lisy2015}.  Partially Observable Monte Carlo Planning (POMCP; \cite{Silver2010}) derives effective strategies by building a search tree of game histories online, but was only tested on much smaller games. Monte Carlo counterfactual regret minimization (CFR)~\cite{Lisy2015} may prove to be the best approach for RBC, but an immediate, practical application is not obvious due to the large numbers of information sets and states within each information set in RBC. For example, by our previous estimate, 7 moves into the game and even with perfect knowledge of the position of her opponent's pieces, a player would have to account for approximately $36^7 \approx 7.8*10^{10}$ possible information sets for the opponent simply due to the possible states of the opponent's knowledge.  To choose an effective strategy using any of the current CFR approaches of which we are aware, one would need to either sample many actions from each of the possible information sets or effectively eliminate the information set as being unlikely.

\subsection{Reduction to Subgames}
A key contributor to algorithm success in games of perfect information
is the ability to reduce search to subgames as the game progresses.
Namely, in games of perfect information, a player can consider the
current state of the game and all states that are reachable from that point, ignoring the past.
 Games of imperfect information,
however, cannot be trivially reduced.  In general, portions of the
game that are unreachable from the current state affect current and
future probabilities of unknown information including the opponents'
strategies.  E.g., an opponent is more likely to be in a state that
may have yielded a high reward in the past, even if the reward is no longer
possible.  Additionally, she will act according to her estimate of the
probability distribution of one's own possible states, including
states that one privately knows are not possible.

Research in games of imperfect information has enabled theoretically
sound decomposition of imperfect information games into
subgames~\cite{imp_decomposition,nested_subgame,Moravcik2017,continual_resolving},
which was also a key piece of the algorithms employed by the
super-human poker systems DeepStack~\cite{Moravcik2017} and
Libratus~\cite{Brown2017}.  However, it is not obvious how to
practically decompose RBC into subgames.
For continual resolving~\cite{continual_resolving}, subgames are
defined be a set of game histories that are closed under
indistinguishability by at least one player (``public
subgames'')~\cite{continual_resolving}.  Largely due to the property
mentioned above, where in general, a player does not know what other
players have learned about her state in RBC, this closure can frequently include
most of the original game.   



\section{Conclusion}
Reconnaissance blind chess provides several unique properties that are not found in other games of which the authors are aware.  It maintains the raw strategic complexity of chess while incorporating significant uncertainty that is greater than that of poker.  Ultimately, the size of the game becomes comparable to that of the number of possible board configurations of Go.  Unlike many commonly studied games which have private and public information, exclusively, RBC involves a large amount of uncertainty about an opponent's knowledge.  This more closely resembles real-world  scenarios where groups are frequently unaware of what their potential adversaries have learned. From an algorithm development perspective, this characteristic of RBC may yield significant challenges by creating an exponential increase in the size of a perceived state.  The complexity and unique properties of reconnaissance blind chess may make it a valuable tool for conducting research into machine decision-making under uncertainty.

\section{Acknowledgments}
We would like to thank Andy Newman and Casey Richardson for their efforts in creating and implementing reconnaissance blind chess.  We would additionally like to thank Cash Costello for his work on the application programming interface (API) and strategies for RBC as well as Corey Lowman, William Li, and Nathan Drenkow for their contributions on bot strategy.

\bibliography{references}
\bibliographystyle{aaai}
\end{document}